\documentclass{article}
\usepackage[utf8]{inputenc}
\usepackage{titling}
\usepackage{graphicx}
\usepackage{placeins}
\usepackage[left=1in,right=1in,top=1in,bottom=1in]{geometry}
\usepackage{amsmath}
\usepackage{amssymb}
\usepackage{xcolor}
\usepackage{url}
\usepackage{hyperref}
\usepackage{booktabs}
\usepackage[normalem]{ulem}
\usepackage{titling}
\usepackage{float}
\usepackage{subcaption}
\usepackage{comment}
\usepackage{authblk}
\usepackage{pdflscape}
\usepackage{afterpage}

\title{Network Embedding Exploration Tool (NEExT)}

\author[1]{Ashkan Dehghan\thanks{email: ashkan.dehghan@torontomu.ca}}
\author[1]{Pawe\l{} Pra\l{}at}
\author[2]{Fran\c{c}ois Th\'eberge}

\affil[1]{Toronto Metropolitan University, Toronto, ON, Canada}
\affil[2]{Tutte Institute for Mathematics and Computing, Ottawa, ON, Canada}

\begin{document}

\maketitle

\begin{abstract}
Many real-world and artificial systems and processes can be represented as graphs. Some examples of such systems include social networks, financial transactions, supply chains, and molecular structures. In many of these cases, one needs to consider a collection of graphs, rather than a single network. This could be a collection of distinct but related graphs, such as different protein structures or graphs resulting from dynamic processes on the same network. Examples of the latter include the evolution of social networks, community-induced graphs, or ego-nets around various nodes. A significant challenge commonly encountered is the absence of ground-truth labels for graphs or nodes, necessitating the use of unsupervised techniques to analyze such systems. Moreover, even when ground-truth labels are available, many existing graph machine learning methods depend on complex deep learning models, complicating model explainability and interpretability. To address some of these challenges, we have introduced \textbf{NEExT\footnote{\url{https://github.com/ashdehghan/NEExT}}} (\textbf{N}etwork \textbf{E}mbedding \textbf{Ex}ploration \textbf{T}ool) for embedding collections of graphs via user-defined node features. The advantages of the framework are twofold: (i) the ability to easily define your own interpretable node-based features in view of the task at hand, and (ii) fast embedding of graphs provided by the \textbf{Vectorizers}\footnote{\url{https://vectorizers.readthedocs.io}} library. In this paper, we demonstrate the usefulness of \textbf{NEExT} on collections of synthetic and real-world graphs. For supervised tasks, we demonstrate that performance in graph classification tasks could be achieved similarly to other state-of-the-art techniques while maintaining model interpretability. Furthermore, our framework can also be used to generate high-quality embeddings in an unsupervised way, where target variables are not available.
\end{abstract}

\section{Introduction}

Complex systems can often be modeled as networks that represent both the individual properties of each component and the relationships between them. Examples of such systems include social networks, financial transaction networks, biological systems such as proteins or ecosystems, and many more. Although such systems are often easy to think about as individual static networks, in reality they often present themselves as collections of networks. This can be either an intrinsic property of the system, such as a collection of protein molecules, or it can be produced as a function of some dynamics. For example, the collection could represent snapshots of a social network evolving over time, or collection of subnetworks created from ego-nets of nodes in the same graph. Representing complex systems as networks or collections of networks is only the first step in the workflow to analyze and study complex systems.

Practitioners in the field of \textit{network science} are often interested in performing predictive type analysis on complex networks. Such analysis could include building classification or regression models of graphs, when ground truth labels are available, or performing analysis using unsupervised techniques, when the labels are not present. In many of such cases, representing graphs as continuous vectors, called embeddings, can enable one to use many traditional machine learning algorithms that take as input structured continuous vectors of integers or floating values. Embedding algorithms can be designed to capture representations at various levels such as node embeddings, edge embeddings, subgraph embeddings, or embeddings of the entire graph. Node embedding is a transformation of the nodes of a network into a set of vectors~\cite{aggarwal2020machine}. Due to their spectacular successes in various applications, they are becoming increasingly popular in the ML community. There are more than 100 algorithms available to use and frameworks to evaluate them (such as~\cite{kaminski2022multi}). Independently, many analytic tasks (such as classification, clustering, and regression) in various domains, including social networks, cybersecurity, bioinformatics, and chemoinformatics, require representing graphs as fixed-length feature vectors~\cite{aggarwal2020machine}. For example, embeddings of graphs representing program calls could be used to detect malware~\cite{narayanan2016subgraph2vec}, embeddings of graphs representing chemical compounds could be used to predict properties of the associated compounds, such as solubility and anticancer activity~\cite{yanardag2015deep,narayanan2017graph2vec}. 

Historically, graph kernels have been considered to be a standard way to handle the above graph analytics tasks. In this approach, the similarity (kernel value) between pairs of graphs is computed by recursively decomposing them into simpler substructures (such as random walks, shortest paths, and graphlets) and defining similarity (kernel) between these substructures. After that, some standard kernel methods, such as Support Vector Machines (SVMs), can be used to classify or cluster graphs. Note that many algorithms of this nature do not explicitly produce graph embeddings, and so they cannot be immediately used for general ML tasks. 

To overcome this limitation, another powerful technique was recently introduced in the literature~\cite{togninalli2019wasserstein,kolouri2021wasserstein}. We start by extracting features of the nodes of a graph $G$ through some node embedding algorithm. This cloud of $n$ points, corresponding to vectors of features of $n$ nodes of $G$, can be easily normalized so that it can be viewed as the probability distribution in a metric space equipped with a distance, such as the Euclidean distance. Then, the Wasserstein distance, known in the literature under many different names including the Monge--Kantorovich--Rubinstein distance, Kantorovich distance, Mallows distance, earth-mover's distance, or optimal transport distance~\cite{sommerfeld2018inference}, can be used to measure the distance between two graphs by computing the distance between the two corresponding probability distributions. The Wasserstein distance is a metric and is linked to the optimal transport problem~\cite{villani2009optimal} which aims to find an optimal way to transport the probability mass associated with one graph to the one associated with another one. As already mentioned, this distance is sometimes referred to as the earth mover's distance since in the 3-dimensional case one can think of it as moving piles of dirt in an optimal way. Finally, some algorithm is used to embed graphs into $k$-dimensional space of vectors such that the Wasserstein distance between graphs matches the distance between the corresponding vectors as much as possible. 

\bigskip

In this paper, we introduce a framework that builds on ideas from~\cite{togninalli2019wasserstein,kolouri2021wasserstein}. In addition to providing an efficient and user-friendly exploratory network analysis tool, our main contribution can be summarized as follows. 
\begin{itemize}
    \item The framework not only utilizes a number of standard classical and structural node embeddings but allows to include hand-crafted, user-defined feature vectors that, for example, measure the distribution of power (by including Pagerank or some other centrality measures) or expansion of ego-nets around nodes. This approach has a few immediate benefits: it is much faster to compute such features than to embed all nodes, the results are interpretable and more robust.
    \item The framework utilizes various techniques and metrics for approximating the distances between graphs such as Wasserstein distance and Sinkhorn algorithms, in addition to a computationally efficient approximate Wasserstein vectorization approach. These tools are available and maintained in the {\tt Vectorizers} package.
    \item For supervised learning (when labels for graphs are available), the framework selects (in an automated way) a subset of available node features for the best outcome of a given ML supervised task at hand such as classification or regression. The explainable nature of our features in addition to this feature selection technique will provide us with information on which features are most important and predictive for a given task.
    \item For unsupervised learning, the framework selects (again, in an automated way) a subset of available node features via feature selection algorithm using the largest variance type of criteria.
    \item The framework is designed to be scalable for large graph datasets by leveraging fast embedding algorithms. In addition, we have integrated graph sampling to further extend the efficiency and scalability of our framework. 
\end{itemize} 

This paper details \textbf{NEExT}'s architecture and showcases its efficiency and versatility in extracting meaningful insights from a collection of complex synthetic and real-world networks. This is demonstrated using a collection of supervised as well as unsupervised experiments. All the experiments can be found within the \textit{experiments} folder in the GitHub repository for the \textbf{NEExT} framework~\footnote{\url{https://github.com/ashdehghan/NEExT}}. The paper is structured as follows:
\begin{itemize}
    \item In Section~\ref{sec:tool}, we outline and explain in detail various components of the \textbf{NEExT} framework. This includes the data pre-processing, node-feature generation, embedding algorithms and various techniques used for computing feature importance.
    \item In Section~\ref{sec:experiments} we use this framework to perform numerous experiments on both synthetic and real-world networks. In the synthetic network experimentation section, we focus mainly on outlining various properties and functionality of the network. Meanwhile, in the real-world experimentation, we focus on highlighting the value of the framework from the perspective of industry practitioner. In addition, we introduce a sampling technique for improving computational efficiency of the framework and show the impact of sampling on embedding quality and downstream machine learning tasks. 
    \item Lastly, in Section~\ref{sec:conclusion}, we provide a summary of our work and highlight some potential future research directions.
\end{itemize}
Finally, we would like to point out that this paper is the extended version of the proceeding paper presented at the International Workshop on Algorithms and Models for the Web-Graph in 2024~\cite{dehghan2024network}.

\section{Network Embedding Exploration Tool (NEExT)}\label{sec:tool}

Suppose that we have a collection of $m$ graphs, $G_1, G_2, \ldots, G_m$, that we wish to analyze. The graphs ($G_i$) in the dataset may not have comparable properties such as degree distributions, densities, structure, etc. In particular, they can have different number of nodes, and they usually do. Our tool creates an embedding of this collection of graphs by assigning to each of the $m$ graphs a $d$-dimensional vector. 

Our approach consists of the following 3 steps (details are provided in the following subsections):
\begin{enumerate}
    \item Pre-process the collection of graphs (Subsection~\ref{subsec:pre-processing}).
    \item For each graph $G_i$ with $n_i$ nodes, build $k$-dimensional vector representations for all the nodes (default option) or for some random subset of the nodes (faster but approximated option) (Subsection~\ref{subsec:vectorizing}). 
    \item Given $m$ families of $k$-dimensional vectors, one family for each graph, compute $d$-dimensional embedding of the graphs using the {\tt Vectorizers} package (Subsection~\ref{subsec:embedding}). 
\end{enumerate}


In the step~2 above, we assumed that a set of $k$ node features is already identified and it is used to create $k$-dimensional vectors. This selection process has flexibility and there are two possible modes the tool can be used in.
\begin{itemize}
    \item [(a)] \emph{Supervised mode} in which features are selected to create a specific embedding, tuned for a given Machine Learning task at hand (Subsection~\ref{subsec:supervised_mode}).
    \item [(b)] \emph{Unsupervised mode} in which features are selected to provide a universal, all-purpose embedding (Subsection~\ref{subsec:unsupervised_mode}).
\end{itemize}

\subsection{Pre-processing}\label{subsec:pre-processing}

In the pre-processing step, \textbf{NEExT} loads each graph into a \textbf{GraphCollection} object, assigning to each graph appropriate details such as graph labels, graph statistics, etc. In this step, we can also do some preprocessing and data cleaning such as making sure that graphs are connected by replacing each graph with its largest connected component or even its $k$-core (for some chosen value of $k$). The \textbf{GraphCollection} provides a unified and convenient way of extracting graph statistics and manipulating graphs. 

\subsection{Vectorizing the Nodes}\label{subsec:vectorizing}

In this step, we compute node level features on each graph. Several methods to obtain vector representations for the nodes of each graph are available in the framework. However, one advantage of \textbf{NEExT} is that it is easy to add other vector representations for the nodes, which can be interpretable and designed specifically for the types of graphs one wants to analyze and down-stream Machine Learning task using the generated graph embedding. In particular, any classical or structural node embedding can be used such as \textbf{Node2Vec} or \textbf{Struct2Vec}. It is also possible to inject some external features associated with the nodes, that could be completely independent from the graph structure. For example, nodes can be associated with users of some social media and external features can encapsulate the text a given user generates on this social platform or the number of likes they gave to other users. Lastly, users could also define custom metrics and algorithms, which can easily be side-loaded into \textbf{NEExT}, and be used to compute node features.

\medskip

Below, we describe the features that are currently available in \textbf{NEExT} that we used for the experiments presented in this paper.

\medskip \noindent 
\textbf{LSME}.
One of the built-in structural embedding algorithms is called \textbf{Local Signature Matrix Embedding (LSME)}\footnote{\url{https://github.com/ashdehghan/LSME}}. This technique uses a random-walk algorithm to capture local structural properties of nodes. The algorithm returns a $k$-dimensional vector for each node $n_i$, where each element measures the transition probability between various neighbourhoods around a given node $n_i$.

\medskip \noindent 
\textbf{Centrality measures}.
We consider various commonly used centrality measures such as \textbf{PageRank}, \textbf{Closeness Centrality}, \textbf{Degree Centrality}, \textbf{Eigenvector Centrality} and \textbf{Load Centrality}. Details for those measures can be found in, for example,~\cite{kaminski2022book}.

\medskip \noindent 
\textbf{Self-Walk}. This algorithm measures the number of walks of length $k$ that start from a given node and end back at the same node. This can be computed by raising the adjacency matrix to the power of $k$ and then extracting the diagonal elements of the matrix. This feature is yet another metric for measuring the structural property of the graph.

\medskip \noindent 
\textbf{Expansion properties}.
This is an example of a hand-crafted node feature that aims to capture expansion property of a graph. For each node $v$, let $\hat{m}_i$ be the number of neighbours at distance $i$ from $v$, for $i \in \{ 1, 2, \ldots, k\}$. The goal is to embed nodes of possibly different degrees that expand in a similar way close to each other.  To that end, we consider the following feature vector for a node $v$:
$$
E(v) = \left( \frac {\hat{m}_1}{1 \cdot \bar{d}}, \frac {\hat{m}_2}{\hat{m}_1 \cdot \bar{d}}, \ldots, \frac {\hat{m}_k}{\hat{m}_{k-1} \cdot \bar{d}} \right),
$$
where $\bar{d} = \frac {1}{N} \sum_{v \in V} \deg(v) = \frac {2 |E|}{N}$ is the average degree. For good expanders, one would get a collection of vectors that are close to $(1,1,\ldots,1)$. 

\bigskip

LSME and expansion naturally produce $k$ dimensional vectors. For other features (such as 1-dimensional centrality measures described above), one can compute them recursively up to some maximum value $k$ (resulting a $k$ dimensional feature vector) by averaging their values for neighbours up to $j$ hops away for $1 \le j \le k$. Moreover, features can be concatenated to obtained even higher dimensional representations, if needed. 

\subsection{Embedding of the Graphs}\label{subsec:embedding}

Recall that at this step of the process we have a collection of graphs $G_i$  ($1 \le i \le m$) with respectively $n_i$ nodes, and a $k$-dimensional vector representation for each node, in each graph. Then, each graph can be seen as a distribution of points over $k$-dimensional space, and we can use some measure of distance between distributions to embed the graphs in some vector space.

One possible candidate is the well-known total variation distance (see, for example,~\cite{van2024random}). Although powerful for many theoretical applications, it has practical limitations. It only considers the maximum disagreement between the measures, and does not take into account the underlying metric space of the measures. Another possible candidate is Kullback-Leibler (KL) divergence, or relative entropy~\cite{kullback1951information}. However, KL-divergence is not symmetric and does not satisfy the triangle inequality. Although it encapsulates a notion of a ``distance'', it is not a metric. In addition, similarly to the total variation distance, it does not take the underlying metric space of the measures into account.

We decided to use the Wasserstein distance, which is obtained by finding the optimal transport plan between distributions. This measure is also known as the earth mover's distance between distributions (i.e.\ measure the amount of ``work'' to move mass from one distribution to the other); see, for example,~\cite{togninalli2019wasserstein}. This distance is symmetric and can be shown to satisfy the triangle inequality~\cite{clement2008elementary}.
Embedding graphs this way is similar to the context of document embedding\footnote{\url{https://vectorizers.readthedocs.io}}, where each word is represented by a vector (obtained via some word embedding algorithm), and each document is a ``bag of word vectors''.

Finding an optimal transport plan to calculate the Wasserstein distance between two discrete distributions can be formulated as a linear program, and solved in polynomial time by the Simplex algorithm~\cite{dantzig1951application}. Unfortunately, given $m$ graphs, computing all such distances requires estimating $\Theta(m^2)$ pairwise distances, which has a high computational cost. One solution to this issue is to define some reference distribution (for example via averaging the vectors), and find the optimal transport plan from each graph's ``bag of vectors'' to this reference distribution. This reduces the number of optimal transport calculations from $\binom{m}{2}$ to $m$. This is know as linear optimal transport (LOT)~\cite{villani2009optimal}, which is used, for example, in~\cite{kolouri2021wasserstein}.
We use the implementation of this approach from the easy to use {\tt Vectorizers}
Python package, which solves the LOT and computes embeddings by computing the SVD (Singular Value Decomposition) of the optimal transport plans. 

Computing the Wasserstein distances, even using a reference distribution, can still be prohibitive for some large problems. 
We therefore consider two faster methods which are also implemented in {\tt Vectorizers}. 
The first one uses the Sinkhorn distance which is based on entropic regularization of the transport plans; see~\cite{cuturi2013sinkhorn}.
The other one, ApproximateWasserstein, also solves the LOT but using a single-point reference distribution obtained via averaging, as described in~\cite{arora2017ASB}. Embeddings are obtained using SVD with scaling according to the singular values.

As a rule of thumb, when $k \ll m$, one can use the Wasserstein or Sinkhorn approach while for larger $k$, the ApproximateWasserstein can be used for better performance.



\subsection{Supervised Mode}\label{subsec:supervised_mode}

There are many different types of node features that one can use to create vectors associated with nodes of a graph in step~2 of the process. The quality of the produced embedding depends on which features are included. Unfortunately, there is no reason to expect that some specific collection of features produces a high-quality embedding for all possible graphs and all possible down-stream tasks. The choice depends on both the family of networks that one wishes to analyze and a given task at hand. In general, including all available features is typically not a good idea.

In the \textbf{supervised} mode, we assume that some of our graphs are labeled. Then, the framework can aim to create a specific embedding using a carefully selected family of node features (of course, without using the labels) with the goal for the embedding to have a strong predictive power of the labels. Since checking all possible subsets of features is impossible in practice, we propose the following two natural algorithms, \textbf{greedy} and \textbf{fast}, that we describe independently below.

\subsubsection*{Greedy Selection of Features}

Recall that in step~2 of the framework, for each graph $G_i$ with $n_i$ nodes, there are $k$-dimensional vector representations for all the nodes of $G_i$. Each of these $k$ dimensions is associated with one of the node features which we denote as $f_1, f_2, \ldots, f_k$.

The main idea behind the first algorithm is to find a permutation of the features for which the order is determined in a greedy fashion by selecting features with the highest incremental contribution to the accuracy (or other cost functions, such as recall or precision) of a classifier that is built using embeddings that, in turn, use the selected features. The algorithm works in the following way:

\begin{enumerate}
    \item Let $F = \{ f_1, f_2, \ldots, f_k\}$ ($F$ is a family of available features), $F_0 = \emptyset$, and $i=0$.
    
    \item There are $i$ features already included in $F_i$ (the subset of features from the previous iteration). Consider $k-i$ subsets of size $i+1$ of features from $F$ that contain $F_i$. In other words, for each feature $f_j \in F \setminus F_i$ (feature that is not included in $F_i$ yet), consider a subset of features $F_i \cup \{f_j\}$ (that is, complement $F_i$, with each of the remaining features, independently). 
    
    \item For each of the $k-i$ subsets of features of size $i+1$, independently generate an embedding using the selected features. By default, the Approximate Wasserstein method is used as the goal at this point is to have a quick way of evaluating different subsets of features, not to generate a high-quality embedding on them. As a result, the dimension of the generated embeddings is always equal to the number of features used, namely, $i+1$.
    
    \item For each of the $k-i$ generated embeddings, we train several XGBoost~\footnote{\url{https://xgboost.readthedocs.io/en/stable/}} classifier models (50 by default, starting from different random seeds).
    The performance of the models is measured using the accuracy metric and for each of the $k-i$ subsets of features $S \subseteq F$ explored, the average accuracy $a(S)$ is computed. Note that for the XGBoost models, data is split into 70/30, where 70\% of the data is used for training and 30\% of the data is used for testing the model.
    
    \item Let $F_{i+1}$ be a subset $S$ of features with the largest associated value of $a(S)$ (out of the $k-i$ subsets tested in the previous step). Note that it might happen that $a(F_{i+1}) < a(F_i)$, that it, the performance might go up or down in comparison to the previous iteration. \\
    (In other words, to summarize steps 2--5, the feature $f_j \in F \setminus F_i$ with the largest incremental accuracy is selected as the most predictive feature and is then added to $F_{i}$ to create $F_{i+1}$.) 
    
    \item If $i+1 < k$, then we increase $i$ and go back to step~2.
    
    \item Otherwise (that is, when $i+1=k$), the process is almost finished. A list of $k$ features is sorted with respect to their incremental accuracy. There are $k$ potential subsets of features that were evaluated during the process. We select the subset $\hat{F}$ with the largest value of the corresponding $a(S)$ and use it to produce the final embedding, this time using more accurate version of the Wasserstein distance.  
\end{enumerate}

We label the above process the \textbf{Greedy} feature selection method, since at every step it selects the feature with highest incremental contribution to the model. 

\subsubsection*{Fast Selection of Features}

Similar to the greedy algorithm, the main purpose of the \textbf{Fast} Selection Method is to find an ordered set of features with highest to lowest predictive power, for a given predictive task. The main challenge with the \textbf{Greedy} method is that it is computationally expensive to iterate through the features and compute their incremental contribution to the classifier. In the \textbf{Fast} method, we bypass the iterative approach and use the feature importance functionality of the classifier (decision tree method), to identify the set of features with the most predictive power. One thing to note is that the main difference between doing feature importance analysis in a regular machine learning task and the technique described here is that in a typical feature importance analysis, the features used by the classifier are the features we care to analyze. However, in our case, the features we are interested are the features computed on the graphs. The graph features are embedded into an embedding space, and the embedding vectors are used to train a model.
\begin{enumerate}
    \item Let $F = \{ f_1, f_2, \ldots, f_k\}$ represent a family of features we have computed for a given dataset.
    \item Each feature $f_i$ is embedded into a single dimensional embedding vector $e_i$, resulting in a set of $E = \{ e_1, e_2, \ldots, e_k\}$ embedding vectors.
    \item We treat the embedding set $E = \{ e_1, e_2, \ldots, e_k\}$ as the features for the machine learning task, and train a \textbf{Random Forest Classifier} using the embeddings as features.
    \item An ordered set of features can then be extracted from the classifier, which then directly maps to the graph features computed on the dataset.
\end{enumerate}
The \textbf{Fast} method allows us to map feature importance from the embedding space directly to the feature computed on the graphs, since each feature is embedded into a single embedding vector, and therefore there is a one-to-one mapping between features computed on the graph and the embedding vectors used as features by the classifier. This allows us to compute the feature importance in one-shot, rather than computing them iteratively as we do using the \textbf{Greedy} method. It is important to note that the \textbf{Greedy} and the \textbf{Fast} method are not equivalent. In the \textbf{Greedy} method, $k$ features are embedded into $k$ dimensional embedding vectors, which means that embeddings have more expressive power. On the other hand, in the \textbf{Fast} method, each feature is mapped onto a single dimensional embedding vector, and the embeddings are then combined, after the embedding are generated. Lastly, the order of the features in the \textbf{Fast} method is taken directly from the output of the \textbf{Random Forest Classifiers} and we do not perform any additional feature sorting (using the \textbf{Greedy} or other methods) afterwards. We do want to note that an alternative approach is to use the \textbf{Fast} method to construct a smaller subset of features and then use the \textbf{Greedy} method on the smaller subset, which would reduce computational complexity.

\subsection{Unsupervised Mode}\label{subsec:unsupervised_mode}

Both the \textbf{Greedy} and the \textbf{Fast} methods assume that we have target or labelled data for the graphs we are analyzing. This is, however, not the case in every scenario. In fact, acquiring ground truth labels for datasets is often challenging, and most real-world datasets are often without labels or are only partially labelled. To tackle scenarios where labels are not available, we have introduced an \textbf{Unsupervised} algorithm for identifying a diverse selection of features with good general predictive power. Since the actual predictive power of a feature (or embedding) will depend on the task at hand, we have to rely on statistical properties of the features to identify potentially predictive features. Here, we outline the general idea behind the \textbf{Unsupervised} algorithm:
\begin{enumerate}
    \item From each graph $G_i$ in the graph collection, we randomly sample $M$ nodes.
    \item We compute the family $F = \{ f_1, f_2, \ldots, f_k \}$ of node-features on the selected $M$ nodes, for each graph $G_i$.
    \item We then compare the node-feature distributions across a random sample of graphs in the graph collection, for each feature $f_i$, and compute \textit{Wasserstein distance} between the node-feature distributions. This allows us to approximate the variability of a given feature across the graphs.
    \item Once the above metrics are computed, we select the feature with the highest variance in \textit{Wasserstein distance} across the graphs. This feature is the most predictive unsupervised feature and will be used as the seed feature. The intuition here is that the values of the selected feature is diverse across graphs, which could be used to build predictive models of whatever label we have for the graphs.
    \item The second feature is selected by identifying the feature with second highest variance, while also considering correlation with any already selected feature. The goal here is to selected the second highly variable feature across graphs, that does not have strong correlation with the first feature.
    \item This process is repeated until all features are sorted in this manner.
\end{enumerate}
The fundamental idea behind the \textbf{Unsupervised} algorithm is similar to that of the Principal Component Analysis (PCA), but instead of transforming the features into a new coordinate system (principal components), we identify features with the highest variance as measured by the Wasserstein distance measured between their distributions. The main objective here is to identify an ordered set of features with highest potential predictive power across diverse set of tasks. One could use this algorithm in scenarios where labelled data is not available or one wishes to create a general-purpose embedding performing reasonably well for diverse ML tasks.

\section{Experiments}\label{sec:experiments}

In this section, we illustrate the use of our framework by experimenting with synthetic as well as real-life networks. The goal is to explore various capabilities of the framework for both unsupervised and supervised applications. In the first subsection, we use the \textit{\textbf{A}rtificial \textbf{B}enchmark for \textbf{C}ommunity \textbf{D}etection} (\textbf{ABCD}) framework~\cite{kaminski2021artificial}, to generate synthetic graphs. The \textbf{ABCD} framework is a random graph model framework with community structure and power-law distribution for both degrees and community sizes. The goal here is to explore and highlight various properties of our framework in a controlled environment and showcase its use from a practitioner's point of view. Therefore, we consider idealized and synthetically generated cases, while considering more real-world scenarios in the following subsection.

\subsection{Synthetic Graphs}

The \textbf{ABCD} model, an alternative approach to the \textbf{LFR} model~\cite{lancichinetti2008benchmark}, allows us to generate random graphs with control over power-law distribution of both node degrees and of community sizes, fraction of outlier nodes, level of noise, and other parameters. We leverage the Julia implementation\footnote{\url{https://github.com/bkamins/ABCDGraphGenerator.jl}} to generate the synthetic graphs used in this subsection, with a Python version\footnote{\url{https://pypi.org/project/abcd-graph}} also available. Moreover, there exists a faster implementation\footnote{\url{https://github.com/tolcz/ABCDeGraphGenerator.jl/}} that uses multiple threads (\textbf{ABCDe})~\cite{kaminski2022abcde} which can be used to generate huge graphs.

Undirected variant of \textbf{LFR} and \textbf{ABCD} produce graphs with comparable properties but \textbf{ABCD}/\textbf{ABCDe} is faster than \textbf{LFR} and can be easily tuned to allow the user to make a smooth transition between the two extremes: pure (disjoint) communities and random graph with no community structure. Moreover, it is easier to analyze theoretically---for example, in~\cite{kaminski2022modularity,barrett2025self} various theoretical asymptotic properties of the \textbf{ABCD} model are investigated including the modularity function and self-similarities of the ground-truth communities. More importantly, the model is extremely flexible and allows to include outliers~\cite{kaminski2023artificial} (\textbf{ABCD+o}) or to generate hypergraphs~\cite{kaminski2023hypergraph} (\textbf{h--ABCD}).

\medskip

To explore various properties of our framework, we designed three experiments on synthetic networks. We discuss various aspects of the framework but here is the main take-home message. In the first experiment we show that the level of noise is captured by graph embeddings using a variety of features, and explore how dimension of node features affect the quality of the generated embeddings. The second experiment verifies that our methodology is able to capture structural properties of graphs and embed similar graphs close to each other regardless of their sizes. Lastly, the third experiment shows that for a given problem at hand (in this case, detecting the number of outliers), we can use our framework to identify most predictive set of features.

The parameters used in each experiment are outlined in Table~\ref{table:abcd_experiment_details_table}. The number of nodes is equal to $n$. The degree distribution follows power-law with exponent $\gamma$, minimum $\delta$ and maximum $\Delta$. The distribution of community sizes follows power-law with exponent $\beta$, minimum $c$ and maximum $C$. The level of noise is controlled by $\xi$ and represents the fraction of edges in the background graph, where the community structure is ignored. Finally, the number of outliers is equal to $o$. Here, (150 x 10) means that 150 graphs were independently generated with 10 outliers in each of them.

\begin{table}[ht]
  \begin{center}
    \begin{tabular}{|c|c|c|c|}
      \hline
      \textbf{Parameter} & \textbf{Experiment 1}& \textbf{Experiment 2}& \textbf{Experiment 3}\\
      \hline
      \hline
      $n$ & 200 & $\{200, 250, \ldots, 400\}$ & 200 \\
      \hline
      $\gamma$ & 3 & 3 & 3 \\
      \hline
      $\delta$ & 5 & 5 & 5 \\
      \hline
      $\Delta$ & 10 & 10 & 10 \\
      \hline
      $\beta$  & 2 & 2 & 2 \\
      \hline
      $c$ & 10 & 10 & 10 \\
      \hline
      $C$ & 20 & 20 & 20 \\
      \hline
      $\xi$ & $\{0.1, 0.101, \ldots, 0.9\}$ & $\{0.1, 0.2, \ldots, 0.5\}$ & 0.2 \\
      \hline
      $o$ & 0 & 0 & (150 x 10) + (150 x 50)\\
      \hline
      \hline
    \end{tabular}
    \caption{\textbf{ABCD}/\textbf{ABCD+o} Synthetic Graph Parameters.}
    \label{table:abcd_experiment_details_table}
  \end{center}
\end{table}

\subsubsection{Experiment 1 -- Varying Level of Noise.}

In the first experiment, we explore the effect of noise that is controlled by parameter $\xi$ in the \textbf{ABCD} model. We designed this experiment to mimic a dynamic property of a graph in which an incremental change in a property of a parameter in the graph results in the change in underlying graph structure. In a real world network, this could resemble the change in the polarization (or the fragmentation) of a network in which the boundaries between communities slowly vanish as the amount of noise increases. A good embedding of graphs representing different snapshots of such evolving network should capture this dynamic. This is what we aim to verify in this experiment. Of course, we have to note that our synthetic experiment is an idealized version of such dynamic system and real-world scenarios will be more involved.

\begin{figure}[ht]
  \centering
  \includegraphics[scale=0.35]{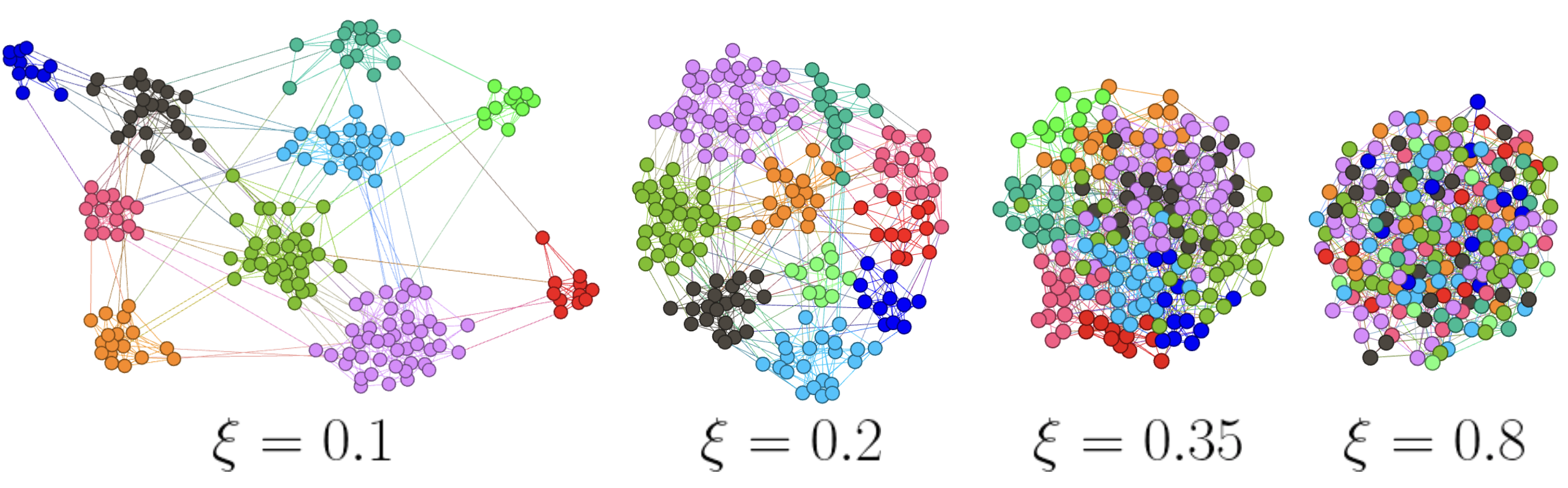}
  \caption{Examples of graphs generated using the \textbf{ABCD} synthetic graph for Experiment~1, as detailed in Table~\ref{table:abcd_experiment_details_table}, for $\xi \in \{0.1, 0.2, 0.35, 0.8\}$. Ground-truth communities are represented with different colours.}
  \label{fig:ABCD_xi_graph_examples}
\end{figure}

We construct 801 graphs, with $\xi$ ranging from $0.1$ to $0.9$ in steps of $0.001$. Recall that in the \textbf{ABCD} model, $\xi$ controls the fraction of edges that fall into the background graph (almost all of these edges are between nodes from different communities). A sample of four graphs from the 801 generated ones are shown in Figure~\ref{fig:ABCD_xi_graph_examples}. Note that these graphs are shown for illustration purposes. We then use $\xi$ values as a label for each graph to be used for a regression task.  

More details will follow but here are high-level steps of this experiment:
\begin{itemize}
    \item Generate a collection of graphs with varying level of noise controlled by $\xi$.
    \item Generate feature vectors of size $k=2$ to $k=8$ for each graph. 
    \item Use some combination of the features vectors to construct graph embeddings of dimension $d=2$.
    \item Use the graph embedding vectors as features for a regression task to predict the $\xi$ values for each graph.
\end{itemize}

Once the graph collection is generated and loaded into our framework, we can compute various graph properties on each graph in the collection. For this experiment, we compute the following graph features: \textbf{Expansion}, \textbf{LSME}, \textbf{PageRank}, \textbf{Closeness Centrality}, \textbf{Degree Centrality}, and \textbf{Eigenvector Centrality}. For each feature, we construct a $k$ dimensional vector. For example, for \textbf{PageRank} as a feature with $k=3$, for each node $v$ we calculate the \textbf{PageRank} value of $v$ as well as the average \textbf{PageRank} values of neighbours at distance $i$ from $v$, where $i \in \{1, 2, \ldots, k-1\}$. We also construct larger feature vectors by concatenating the vectors from multiple features. For example, a feature vector of \textbf{Expansion} + \textbf{LSME} with $k=3$ is a concatenation of a 3 dimensional \textbf{LSME} feature vector and 3 dimensional \textbf{Expansion} feature vector, resulting in a 6 dimensional global feature vector. In total, 8 different combinations of features were selected for experiments.

Having defined the feature generation process, we construct features of lengths $k$ from 2 to 8 based on the above list. We then use the approximate Wasserstein technique to embed each graph in the collection into a two dimensional embedding. We chose embedding dimension $d=2$, since the approximate technique has an upper limit of $k$ for the dimension of the embedded space and the smallest feature vector size is of dimension $k=2$. Moreover, since our graph embeddings are used in a downstream supervised regression task, we wanted to keep the dimensionality of the embeddings the same to standardize the comparison of the models.

\begin{figure}[ht]
  \centering
  \includegraphics[scale=0.5]{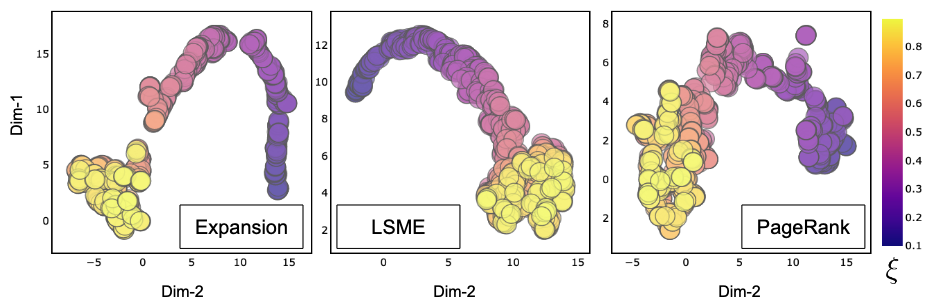}
  \caption{Two dimensional graph embeddings built using the approximate Wasserstein technique and graph features built using the \textbf{Expansion}, \textbf{LSME}, and \textbf{PageRank} node features. The dimension for all the above node embeddings is set to $5$.}
  \label{fig:ABCD_xi_emb_dim_5_expansion_lsme_pagerank}
\end{figure}

In Figure~\ref{fig:ABCD_xi_emb_dim_5_expansion_lsme_pagerank} we show the three 2-dimensional embeddings of graphs that were built using the \textbf{Expansion}, \textbf{LSME} and, respectively, \textbf{PageRank} features. In all three cases, the underlying feature vectors have length $k=5$. Each data point (graph embedding) is then coloured based on the level of noise (that is, parameter $\xi$ used in the \textbf{ABCD} model). It is clear that in all three cases there is a relation between the value of $\xi$ and the graph embedding vector. To explore this relationship further, we use a regression model for graph embeddings that are built using various graph features.

\begin{figure}[ht]
  \centering
  \includegraphics[scale=0.45]{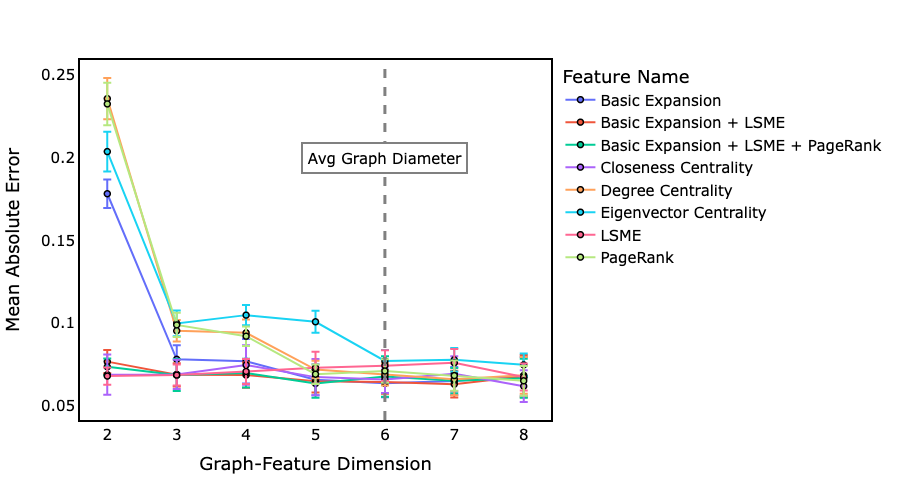}
  \caption{Mean-absolute-error measured for a regression model built to predict $\xi$ in Experiment~1, as defined in Table~\ref{table:abcd_experiment_details_table}. The $x$-axis is the length of the feature vectors computed on each graph. The final graph embedding is uniformly set to $d=2$.}
  \label{fig:ABCD_xi_Experiment_Regression_Model_MAE}
\end{figure}

Using the two dimensional embeddings of the graphs, we train regression models using XGBoost to predict the value of $\xi$ for the unlabeled graphs. In our experiment, the train/test split is set to 70/30 and we repeat each experiment 100 times to arrive at the average mean-absolute-error and the standard deviation over the runs, shown as error bars in Figure~\ref{fig:ABCD_xi_Experiment_Regression_Model_MAE}. Here, the $x$-axis corresponds to $k$, the length of feature vectors computed for nodes of each graph, and different colours correspond to combination of various types of features. 

We start by highlighting the fact that the overall performance of the models increases (the mean-absolute-error decreases) as the length of the underlying feature vectors increases. This is expected, since increase in $k$ corresponds to a larger window for capturing structural properties of the underlying graph. We show that this trend continues until the length of the feature vectors reaches the diameter of the graphs. At this length scale, the feature vectors are capturing global structural properties of the graph. We note that the diameter for our synthetic graphs is relatively small, since we consider graphs of size $n=200$ that are relatively good expanders. Moreover, we note that models built on combination of features perform the best, since each feature captures a different structural property of the graph.

Finally, let us note that in this experiment we were not fine-tuning the XGBoost models to achieve the best performance, but rather to illustrate the predictive power of various graph features and associated graph embeddings. Later in this paper we focus on using our framework to determine feature with most predictive power. To get an embedding with a strong predictive power, one should use one of the built-in supervised modes, using greedy or fast selection of features.

\subsubsection{Experiment 2 -- Varying Network Size.}\label{ssec:exp_synthetic_exp2}

In the second experiment, we explore the ability of our framework to capture structural similarities in collections of graphs. In real systems, it is often important to identify structurally similar graphs, regardless of the size of the network. This is often seen in self-similar systems, such as social networks, where particular property presents itself at different scales~\cite{song2005self,barrett2025self}. To study this effect, we use the \textbf{ABCD} model to generate structurally similar networks of various sizes. We achieve this goal by tuning two parameters: the level of noise ($\xi$) and the number of nodes in each graph ($n$).  As highlighted in Table~\ref{table:abcd_experiment_details_table}, we build a collection of 25 graphs with $\xi \in \{0.1, 0.2, 0.3, 0.4, 0.5\}$ and $n \in \{200, 250, 300, 350, 400\}$. As it was done in Experiment~1, we compute node features for each graph and use the approximate Wasserstein technique to build an embedding vector for each graph. We fix the feature vector size to $k=4$ and graph embedding size to $d=4$. To visualize the final embeddings, we use \textbf{UMAP}\footnote{\url{https://umap-learn.readthedocs.io/en/latest/}}~\cite{mcinnes2018umap} to map the final graph embedding into two dimensions.

\begin{figure}[ht]
  \centering
  \includegraphics[scale=0.22]{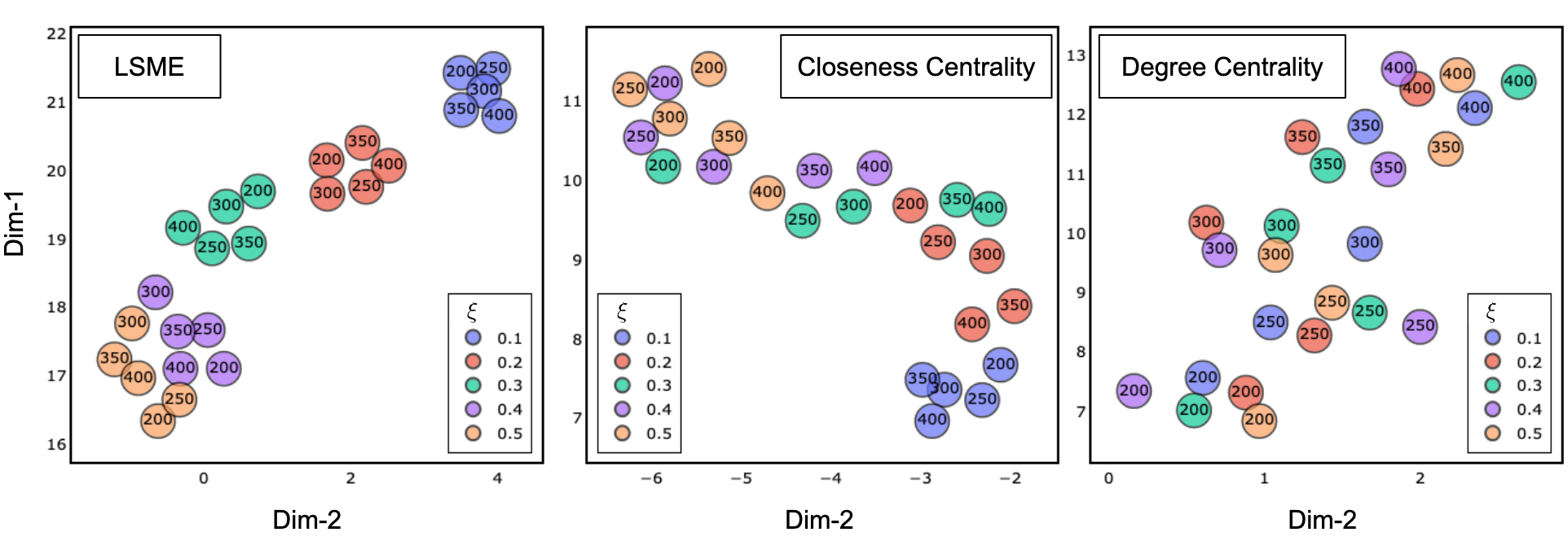}
  \caption{Two dimensional representations of the approximate Wasserstein graph embeddings built using \textbf{LSME}, \textbf{Closeness Centrality}, and \textbf{Degree Centrality} graph features. Colours correspond to different values of noise ($\xi$) and the size of the underlying graphs ($n$) are shown inside each data point.}
  \label{fig:abcd_number_of_nodes_xi_clusters}
\end{figure}

In Figure~\ref{fig:abcd_number_of_nodes_xi_clusters}, we show the two dimensional representations of the graph embeddings, coloured and annotated using $\xi$ and $n$, respectively. In the first chart (left), the underlying feature vector was computed using the \textbf{LSME} structural embedding algorithm. This technique captures structural properties of nodes within each graph. Using \textbf{LSME} as features, the graph embeddings captures similarities between structural properties of each graph. This can be seen in the final two dimension representation of the embeddings, since graphs with similar structure ($\xi$ value, colour-coded) are clustered together. It is interesting to observe that the quality of the clusters (tightness) decreases as the noise ($\xi$) increases.

Similar behaviour is also captured by a more simple structural feature, \textbf{Closeness Centrality}. We can see in the middle chart in Figure~\ref{fig:abcd_number_of_nodes_xi_clusters} that \textbf{Closeness Centrality} also groups graphs with similar properties together, but with lower quality compared to \textbf{LSME}. The decrease in clustering quality as a function of the level of noise is more evident in this case. The important observation in these cases points at the fact that the approximate Wasserstein technique using \textbf{LSME} or \textbf{Closeness Centrality} preserves structural properties of the embedded graphs such as level of noise.

On the other hand, embeddings using {\bf Degree Centrality} tend to group together graphs of similar sizes (see the right chart in Figure~\ref{fig:abcd_number_of_nodes_xi_clusters}). This shows, not surprisingly, that different combinations of node features affect the generated graph embeddings. This is a good and desired property for various reasons. First, in a supervised setting, it allows us to select the right combination of features depending on the problem at hand. As a result, the framework is able to not only create good quality embeddings but it also sheds a light on which node features have the most predictive power. This distinguishes it from many other embedding methods that are not explainable. Finally, in an unsupervised scenario, the framework can select a diverse group of features and create good ``all-purpose'' embeddings.

\subsubsection{Experiment 3 -- Outlier Detection.}\label{ssec:exp_synthetic_exp3}

In this experiment, we control the fraction of outlier nodes in the graph where an outlier is a node that does not belong to any community. We construct two sets of graphs. In the first one, $5\%$ of nodes are outliers and in the second group there are more outliers, namely, $25\%$. For each group, we generate 150 graphs. It is important to note that, since the \textbf{ABCD+o} model is a \textit{randomized} graph generator, each graph in their respective groups are different due to a random nature of the model. This is achieved by setting different random seeds while generating the graphs. 

We compute various graph features on each subgraph and use them to construct graph embeddings using the approximate Wasserstein techniques. Here, we consider 9 different combinations of features, as defined in Table~\ref{table:abcd_outlier_classification_model_details}. Each feature type has a dimension of $k=4$ and we combine different features by concatenating their feature vectors. We note that in this experiment, we train binary classifiers using XGBoost classifier with 70/30 train/test split, and repeat each experiment 100 times to collect enough statistics for model performance.

\begin{table}[h!]
  \begin{center}
    \begin{tabular}{|c|p{10.5cm}|}
      \hline
      \textbf{Model} & \textbf{Feature(s) used} \\
      \hline
      \hline
      M-0 & \textbf{Expansion}  \\ 
      \hline
      M-1 & \textbf{LSME}  \\ 
      \hline
      M-2 & \textbf{PageRank}  \\ 
      \hline
      M-3 & \textbf{Degree Centrality}  \\ 
      \hline
      M-4 & \textbf{Closeness Centrality} \\ 
      \hline
      M-5 & \textbf{Eigenvector Centrality}  \\ 
      \hline
      M-6 & \textbf{Expansion}, \textbf{LSME} \\ 
      \hline
      M-7 & \textbf{Expansion}, \textbf{LSME}, \textbf{PageRank}  \\ 
      \hline
      M-8 & \textbf{Expansion}, \textbf{LSME}, \textbf{PageRank}, \textbf{Degree Centrality}, \textbf{Closeness Centrality}, \textbf{Eigenvector Centrality} \\ 
      \hline
      \hline
    \end{tabular}
    \caption{\textbf{ABCD+o} outlier classification model.}
    \label{table:abcd_outlier_classification_model_details}
  \end{center}
\end{table}

\begin{figure}[ht]
  \centering
  \includegraphics[scale=0.32]{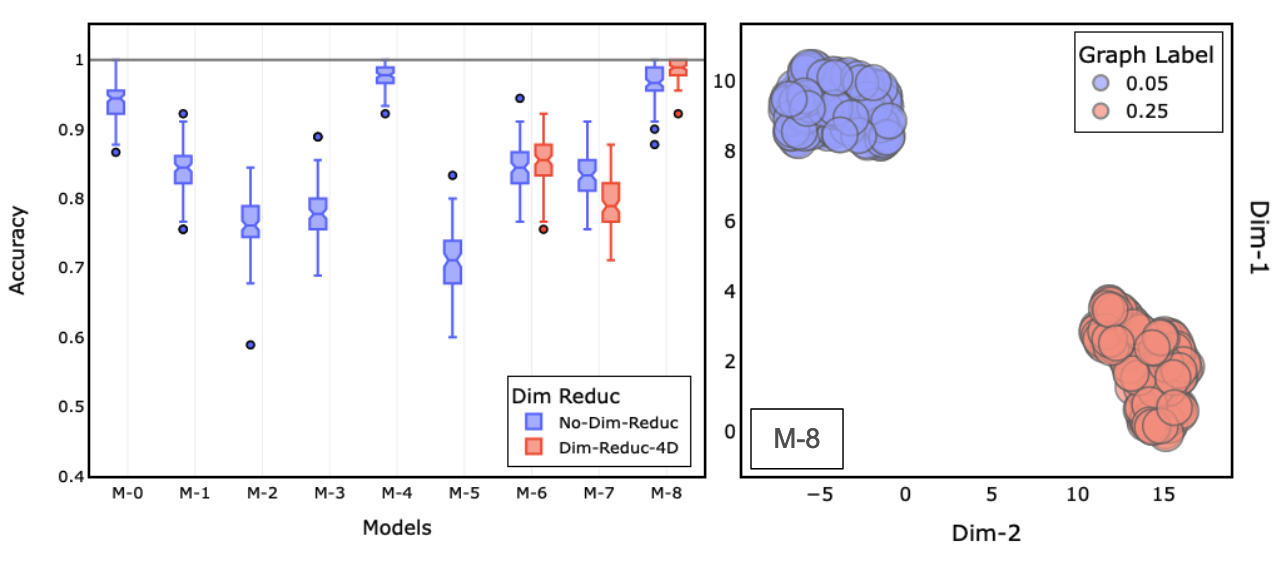}
  \caption{Left: Accuracy of binary-classifiers built for models M-0 to M-8. Right: two dimensional representation of graph embedding vectors built using features in M-8.}
  \label{fig:ABCD_outlier_classifier_accuracy}
\end{figure}

We analyze the performance of binary classifiers (measured using accuracy) trained on graph embedding vectors built using different combinations of feature vectors (M-0 to M-8 as in Table~\ref{table:abcd_outlier_classification_model_details}). In Figure~\ref{fig:ABCD_outlier_classifier_accuracy} (left), we show the model accuracy for each feature set. Focusing on single feature models (M-0 to M-5), we can see that graph embeddings built on top of \textbf{Closeness Centrality} (M-4) perform the best, while models built using \textbf{Eigenvector Centrality} (M-5) perform poorly. It is also worth mentioning that \textbf{Expansion} (M-0), an easy and fast to compute node feature, does very well. The predictive power of \textbf{Closeness Centrality} as a node feature comes from the fact that outlier nodes are, on average, closer to other nodes in the network, since they do not belong to any of the communities but rather randomly connected to the entire network. Therefore, this can be a distinctive factor for graphs with higher number of outlier nodes.

Additionally, we consider composite models, where embeddings are generated from a combination of feature vectors. In Figure~\ref{fig:ABCD_outlier_classifier_accuracy} (left), models M-6, M-7, and M-8 are composite models (as defined in Table~\ref{table:abcd_outlier_classification_model_details}). For each one these models, we run two sets of experiments. One in which we do not apply any dimensionality reduction to the composite feature vectors, before passing them to the graph embedding layer. And, another where we apply dimensionality reduction using \textbf{UMAP}, to reduce the dimension of the composite feature vector to $k=4$. We can see that in all three cases, dimensionality reduction does not have a significant effect on the performance of the models. One thing worth highlighting is that the composite model (M-8) built using all features with $k=4$ performs the best. This hints at the fact that one could capture a wide variety of features in the feature computation layer, then reducing the feature space using a technique such as \textbf{UMAP} to allow for a better performance of a given machine learning model at hand.

Finally, in Figure~\ref{fig:ABCD_outlier_classifier_accuracy} (right) we show a two dimensional clustering of the graph embeddings built using M-8 features. It is clear that the two dimensional representations of graph embedding vectors form two well separated clusters. In an unsupervised setting, one could use a technique such as \textbf{DBSCAN}~\cite{ester1996density} to identify these two clusters, even if the underlying classes are not known. In this experiment however, we know that the underlying graphs are generated using random graph technique with two outlier settings. We show the fraction of outliers for each group as different colours in this figure.

\subsection{Real-World Networks}

In this section, we explore the performance of our framework on a few collections of real-world networks. These collections were acquired from the Benchmark Data Set provided by the department of computer science of TU Dortmund\footnote{\url{https://ls11-www.cs.tu-dortmund.de/staff/morris/graphkerneldatasets}}. A summary of these collections of networks used for our experiments is provided in Table~\ref{table:real_world_networks_table}. Here, we consider five real-world collections; \textbf{IMDB}~\cite{yanardag2015deep}, \textbf{MUTAG}~\cite{debnath1991structure}, \textbf{NCI1}~\cite{wale2008comparison}, \textbf{BZR}~\cite{sutherland2003spline}, and \textbf{PROTEINS}~\cite{borgwardt2005protein}, with various sizes and source, each of which consists of networks that can be categorized into two classes. Therefore, a natural type of analysis would be to investigate the performance of binary-classifiers trained on graph embeddings generated by \textbf{NEExT}. 

\begin{table}[h!]
  \begin{center}
    \begin{tabular}{|c|c|c|c|}
      \hline
      \textbf{Name} & \textbf{\# of Graphs}& \textbf{\# of Classes}& \textbf{Avg.\ \# of Nodes/Edges}\\
      \hline
      \hline
      \textbf{IMDB} & 1000 & 2 &  19.77/96.53  \\
      \hline
      \textbf{MUTAG} & 188 & 2 &  17.93/19.79  \\
      \hline
      \textbf{NCI1} & 4110 & 2 &  29.87/32.30  \\
      \hline
      \textbf{BZR} & 405 & 2 & 35.75/38.36 \\
      \hline
      \textbf{PROTEINS} & 1113 & 2 & 39.06/72.82 \\       
      \hline
      \hline
    \end{tabular}
    \caption{Summary of real-world networks used in our experiments.}
    \label{table:real_world_networks_table}
  \end{center}
\end{table}

In our analysis, we first use the performance of publicly available models as a benchmark for comparing \textit{out-of-the-box} \textbf{NEExT} to other techniques. Our experiments seem to be promising. The main conclusion is that the accuracy of models built using \textbf{NEExT} is similar to other models, even without performing any fine-tuning.

One of the main requirements for performing effective data-science is the task of exploring the importance and contribution of various features in a predictive task. To that end, three feature exploration capabilities of \textbf{NEExT} were introduced. Below, we showcase the two supervised modes (greedy and fast selection of features) and the unsupervised one.

\subsubsection{Experiment 4 -- Comparison to Other Techniques}

In this exploratory stage of our project, the goal is not to seek the best performing model at all cost, but rather provide a framework that can easily create models with reasonable performance compared to state-of-the-art techniques, while keeping model explainability. Moreover, note that some models are trained on additional metadata available for nodes as well as edges. We do not do it at present but it would be easy to incorporate such additional information in our model. It is expected that after appropriate selection of node features and fine-tuning the model, the accuracy of the corresponding models should increase.

In our experiments on the real-networks listed in Table~\ref{table:real_world_networks_table}, we build $d=24$ dimensional graph embeddings using the approximate Wasserstein technique on top of $k=24$ dimensional node feature vectors computed on each network. Here, we use $4$-dimensional concatenated \textbf{LSME}, \textbf{Expansion}, \textbf{Degree Centrality}, \textbf{Closeness Centrality}, \textbf{Load Centrality}, and \textbf{Eigenvector Centrality} as our node feature vectors. We then train the XGBoost binary classifier on top of the $d=24$ dimensional graph embedding feature vectors. Similarly to the approach taken before, we keep a 70/30 train/test split, and repeat each experiment 100 times to build the statistics for our model performance. Lastly, we point out that we do not balance the datasets in our models, to allow the classifiers to capture statistical imbalances in the underlying data distribution.

\begin{figure}[ht]
  \centering
  \includegraphics[scale=0.43]{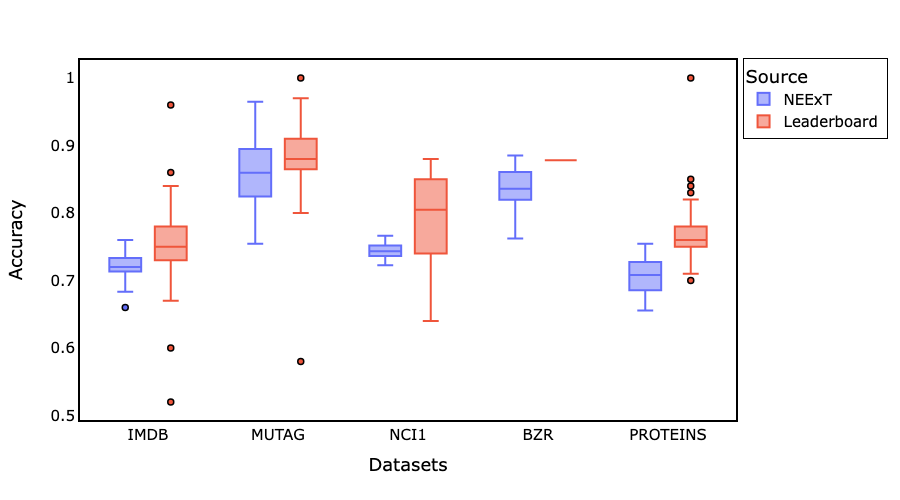}
  \caption{
  Accuracy of models built using \textbf{NEExT} framework (blue) and publicly available models (red), for various real-world networks. The performance of other models is collected from leaderboard chart available on-line.
  }
  \label{fig:neext_real_network_accuracy_chart}
\end{figure}

\begin{table}[ht]
  \begin{center}
    \begin{tabular}{|c|c|c|c|c|c|}
      \hline
      \textbf{Name} & \textbf{Accuracy}& \textbf{Precision} & \textbf{Recall} & \textbf{F1-Score} & \textbf{LB Accuracy}\\
      \hline
      \hline
      \textbf{IMDB} & $0.72 \pm 0.02$ & $0.72 \pm 0.03$ & $0.72 \pm 0.04$ & $0.72 \pm 0.02$ & 0.52--0.96 \\
      \hline
      \textbf{MUTAG} & $0.85 \pm 0.05$ & $0.84 \pm 0.08$ & $0.82 \pm 0.08$ & $0.83 \pm 0.08$ & 0.58--1.00 \\
      \hline
      \textbf{NCI1} & $0.74 \pm 0.01$ & $0.75 \pm 0.02$ & $0.75 \pm 0.01$ & $0.74 \pm 0.01$ & 0.64--0.88 \\
      \hline
      \textbf{BZR} & $0.84 \pm 0.03$ & $0.77 \pm 0.03$ & $0.71 \pm 0.02$ & $0.73 \pm 0.02$ & $0.87$ \\
      \hline
      \textbf{PROTEINS} & $0.71 \pm 0.02$ & $0.70 \pm 0.05$ & $0.69 \pm 0.04$ & $0.69 \pm 0.03$ & 0.70--0.85 \\       
      \hline
      \hline
    \end{tabular}
    \caption{Summary of Real-World Networks Classification Results. Here, \textbf{LB} Accuracy refers to the range of accuracy of the models from the leaderboard.}
    \label{table:real_world_networks_classification_results}
  \end{center}
\end{table}

In Figure~\ref{fig:neext_real_network_accuracy_chart} and Table~\ref{table:real_world_networks_classification_results}, we show the performance of classifiers trained using \textbf{NEExT} and benchmark models collected from leaderboard (\textbf{LB}) chart available on-line\footnote{\url{https://paperswithcode.com/task/graph-classification}}. Note that the \textbf{LB} accuracies are shown as the range of accuracies from various models submitted for each dataset. We see that the accuracy of models built using \textbf{NEExT} is similar to other models.

\subsection{Feature Analysis and Exploration}

In the previous sections, we used our framework to explore and analyze some predictive modelling use-cases using synthetic and real-world graphs. We focused on highlighting some general capabilities of our framework, without specific focus on identifying the most predictive features or optimizing for model performance. In this section, following a data-science and machine learning approach, we focus on feature exploration and optimization. The main idea here is to identify an ordered list of features, for which the order is defined by the highest incremental contribution to the accuracy of classifier, built using embeddings of the features.

In Section~\ref{subsec:supervised_mode}, we introduced three methods, two supervised and one unsupervised, for performing feature importance selection. These method are labelled \textbf{Greedy}, \textbf{Fast}, and \textbf{Unsupervised}. Here, we perform a number experiments, using these techniques to identify the most predictive feature for a given dataset. To provide a reference, we also include \textbf{Random} and \textbf{Worst} case scenarios for performing feature selection. The \textbf{Worst} case scenario is determined by running the \textbf{Greedy} algorithm but where features are selected with the lowest incremental contribution to the accuracy of the model. As we see this, in a real-word setting, a data science practitioner is faced with the following choices when performing feature selection on a new dataset:

\begin{itemize}
    \item The practitioner could select a \textbf{random} set of features computed on a graph dataset, and build a production model using those features.
    \item The practitioner could select and compute \textbf{all} available features.
    \item The practitioner could select a \textbf{subset} of all features with most predictive power.
\end{itemize}

Of course each approach has its pros and cons when it comes to feature selection. A \textbf{random} selection of feature could be a great first approach for quick analysis of the dataset, however it will most likely result in poor performing model, as there is a chance that the random selection results in the selection of the \textit{least} predictive feature set, which we call \textbf{worst} case scenario. On the other hand, one could compute \textbf{all} \textit{available} features, and build a model using such feature set. This has two drawbacks. Firstly, it is computationally expensive to compute all features on a large dataset in a production setting. Second, adding more features is not always beneficial to the performance of machine learning models, specially due to the curse of dimensionality~\cite{balestriero2021learning}. The ideal situation is to include a \textbf{subset} of features with the most predictive power, which provides a balance between model performance and computational cost. However, this approach has one major drawback, which is the challenge of identifying such subset. This section is devoted to exploration of these scenarios. To focus our attention, we select one dataset from the real-world networks \textbf{NCI1} to run our experiments on. In the following sub-section, we highlight the results of feature selection experiments.

\subsubsection*{Feature Selection Experiments}

To explore the effectiveness of feature selection algorithms of \textbf{NEExT}, we compare \textbf{NEExT}'s \textbf{Greedy}, \textbf{Fast} and \textbf{Unsupervised} algorithms against \textbf{Random} and \textbf{Worst} case feature selection scenarios. All of the above approaches are measured with reference to the case of using \textbf{all} features, in a model building exercise. All the experiments are run in the following manner:
\begin{itemize}
    \item In the first iteration of the experiment, each algorithm selects \textbf{one} feature. The \textbf{Greedy}, \textbf{Fast}, \textbf{Unsupervised} and \textbf{Worst} algorithms select this feature based on the approach described in the previous section. In the \textbf{Random} approach, as the name suggests, this feature is selected at random.
    \item The selected feature is computed on all graphs and embeddings are generated according to the given algorithm.
    \item We build 50 binary classifier using XGBoost algorithm, and the average and standard deviation of the accuracy is computed over the runs.
    \item We select the second feature, according to each algorithm and repeat the process until all features are selected.
    \item Lastly, for the \textbf{Random} algorithm there is an outer loop which repeats the above process 500 times. This is done since in each run the features are selected at random and the total number of features is small. As a result, with non-negligible probability, a random selection could turn out to be good (or bad). Hence, for the \textbf{Random} selection algorithm, the mean and standard deviation are computed based on this outer loop to capture the expected performance of a typical subset of features of a given size. 
\end{itemize}

\begin{figure}[ht]
  \centering
  \includegraphics[scale=0.55]{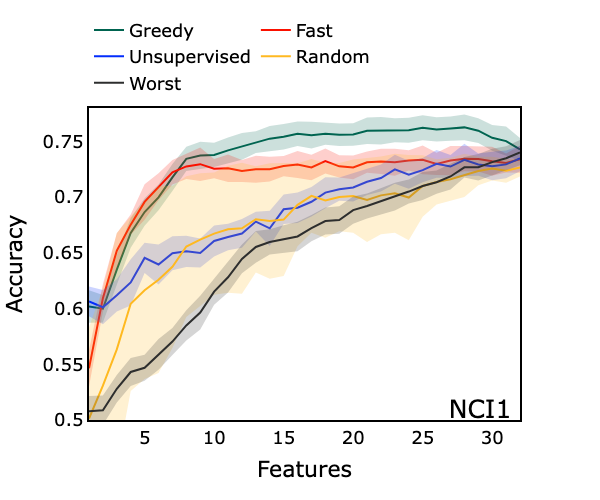}
  \caption{This figure shows the mean (solid line) and standard deviation (shaded area) of model accuracy for the \textbf{Greedy}, \textbf{Fast}, \textbf{Unsupervised}, \textbf{Random}, and \textbf{Worst} algorithms. The $x$-axis is the number features used in each model. For example, $x=5$ means that 5 node-features were used to build embeddings that are used in the binary classifier model.}
  \label{fig:NCI1_all_feature_selection_experiments}
\end{figure}
The results of the above procedure are shown in Figure~\ref{fig:NCI1_all_feature_selection_experiments}, where the mean (solid lines) and standard deviation (shaded areas) are presented for the accuracy of the binary classifiers built for each algorithm. The $x$-axis in Figure~\ref{fig:NCI1_all_feature_selection_experiments} shows the number of node-features used for building the graph embeddings that feed into the classifier models. For all algorithms, the performance of the classifier model often increases as we increase the number of features. This is expected, since adding more features generally leads to the model having more information for determining a classification boundary line. It is important to note that increase in performance of model, as the number of features increase, is not always monotonic and one has to always be aware of the curse of dimensionality. In most cases, there will be an optimal number of features, which depend on many factors such as model capacity, feature diversity and complexity of the problem, just to name a few. 

\medskip

The best performing algorithm is the \textbf{Greedy} method, shown in green. This outcome is consistent with the desired outcome of a greedy technique. At each step, the \textbf{Greedy} method selects the most predictive feature based on the incremental increase to the accuracy of the model. Although the \textbf{Greedy} method produces the best performance, it is computationally expensive, since at each step, it searches across all remaining features for the best performing feature. It is important to note that the \textbf{Greedy} method uses only an exploitative method and may not select the absolute best set of features. A more robust algorithm could mix both exploitation and exploration to determine the most predictive set of features. This will the explored in future studies.

The second best performing algorithm is the \textbf{Fast} method, shown in red. This algorithm performs the feature importance analysis in one step. The \textbf{Fast} method uses the \textit{feature importance} functionality of the \textbf{Random Forest Classifier} to determine the feature importance ranking link\footnote{\href{https://scikit-learn.org/stable/auto_examples/ensemble/plot_forest_importances.html}{Scikit-Learn RandomForestClassifier Feature Importance}}. As described previously, the \textbf{Fast} method, builds one dimensional embeddings of every node-feature, to preserve the one-to-one mapping between node-features and embedding dimensions. Therefore, the output feature importance of the classifier, which is an ordered set of embedding vectors, can be translated directly to the node-features. Although this method has the advantage of computing feature importance efficiently, it also has some disadvantages. The main drawback is that the requirement to preserve a one-to-one mapping between node-features and embedding vectors forces us to map feature into a single dimensional embeddings. This restricts the expresibility of the embedding algorithm, and would result in loss of predictive power, as shown in Figure~\ref{fig:NCI1_all_feature_selection_experiments}. One major benefit of using the \textbf{Fast} algorithm is that it provides you with a quick and approximate method for learning about the predictive quality if the features, and if adding new features would improve your model or not. As we can see, the accuracy of the model plateaus after about 10 features. We can use this information to guide the more expensive \textbf{Greedy} feature search.

Both the \textbf{Greedy} and the \textbf{Fast} methods assume that we have \textit{target} or \textit{labels} for our graphs. This may not always be the case, as acquiring ground truth labels in a real-world datasets can be a challenge. However, one may still be interested in computing embeddings on graph datasets with good general predictive power for unsupervised tasks such as clustering and outlier detection. In this case, you can not rely on the guidance of improvement to accuracy for selecting features. There are two general options, one to select features at random and second to use an unsupervised technique to guide the feature selection.

\medskip

To perform feature importance selection in scenarios where labeled data is not available, we have implemented an unsupervised algorithm, details of which are described in Section~\ref{subsec:unsupervised_mode}. The main motivation behind this algorithm is to provide a selection of diverse set of features that could be used in unsupervised and potentially supervised tasks. To showcase the quality of these features, we compare the performance of the ordered selection of features identified using our \textbf{Unsupervised} algorithm, in a supervised task, against the \textbf{Greedy}, \textbf{Fast} and the \textbf{Random} algorithms. The results of this experiment are presented in Figure~\ref{fig:NCI1_all_feature_selection_experiments}, where the \textbf{Unsupervised} algorithm is shown in blue. Of course, as expected, the \textbf{Unsupervised} algorithm under-performs the more targeted \textbf{Fast} and \textbf{Greedy} methods. Indeed, supervised methods are designed to use the labels to identify the most predictive set of features for a given task, whereas the unsupervised method aims to select a diverse set of general-purpose predictive features. The more interesting comparison is the comparison between the \textbf{Unsupervised} and \textbf{Random} techniques. Randomly selecting features, could result in widely different outcomes. To provide a frame of reference, we plot the average performance of 500 simulations of randomly selecting features and training a classifier on them. The solid yellow line in Figure~\ref{fig:NCI1_all_feature_selection_experiments} shows the mean accuracy of such a simulation. We can see that for the initial 1-10 features, the \textbf{Random} approach greatly under-performs the \textbf{Unsupervised} and naturally the supervised (\textbf{Greedy} and \textbf{Fast}) techniques. Of course, in real-world scenarios, a practitioner will not run hundreds of simulations and may randomly select a set of features. This approach, due to chance, could result in the selection of the set of features with the worst performance, shown in black and labeled \textbf{Worst} in Figure~\ref{fig:NCI1_all_feature_selection_experiments}. Therefore, in situations where labeled data is not available, the \textbf{Unsupervised} algorithm can be used as a good general purpose feature selection technique and will result in a set of diverse features with good predictive power.

In Tables~\ref{table:real_world_greedy_method_node_selection} and~\ref{table:real_world_fast_method_node_selection}, we show the top five most predictive features as determined by the \textbf{Greedy} and \textbf{Fast} methods. As we can see, there are some common features between the two methods. However, as we noted above, the difference in how features are combined and embeddings are generated means that the list of most predictive features will not be the same between different techniques.

\begin{table}[ht]
  \begin{center}
    \begin{tabular}{|l|p{0.6\linewidth}|}
    \hline
    \textbf{Datasets} & \textbf{Top 5 Features} \\
    \hline
    BZR & Page Rank 1, Self Walk 2, Lsme 4, Closeness Centrality 3, Eigenvector Centrality 4 \\
    \hline
    IMDB & Self Walk 1, Eigenvector Centrality 3, Self Walk 2, Load Centrality 2, Load Centrality 4 \\
    \hline
    MUTAG & Page Rank 1, Page Rank 4, Self Walk 3, Load Centrality 3, Lsme 4 \\
    \hline
    NCI1 & Page Rank 1, Load Centrality 3, Load Centrality 4, Self Walk 4, Load Centrality 2 \\
    \hline
    PROTEINS & Page Rank 1, Basic Expansion 2, Lsme 4, Degree Centrality 1, Eigenvector Centrality 2 \\
    \hline
    \end{tabular}
    \caption{List of top most predictive features as selected by the \textbf{Greedy} method.}
    \label{table:real_world_greedy_method_node_selection}
  \end{center}
\end{table}

\begin{table}[ht]
  \begin{center}
    \begin{tabular}{|l|p{0.6\linewidth}|}
    \hline
    \textbf{Datasets} & \textbf{Top 5 Features} \\
    \hline
    BZR & Self Walk 1, Closeness Centrality 3, Lsme 4, Basic Expansion 2, Load Centrality 2 \\
    \hline
    IMDB & Lsme 1, Lsme 2, Self Walk 1, Lsme 3, Lsme 4\\
    \hline
    MUTAG & Eigenvector Centrality 2, Self Walk 1, Eigenvector Centrality 3, Page Rank 4, Page Rank 3  \\
    \hline
    NCI1 & Load Centrality 3, Basic Expansion 3, Load Centrality 4, Eigenvector Centrality 4, Load Centrality 2 \\
    \hline
    PROTEINS & Eigenvector Centrality 1, Closeness Centrality 1, Page Rank 1, Degree Centrality 1, Page Rank 2 \\
    \hline
    \end{tabular}
    \caption{List of top most predictive features as selected by the \textbf{Fast} method.}
    \label{table:real_world_fast_method_node_selection}
  \end{center}
\end{table}

\subsection{Sampling}\label{sec:sampling}

One of the main challenges with applying network science techniques to real-life networks such as social networks, is the computational complexity arising due to the large size of such systems. For example, social networks such as Facebook have billions of users with tens of billions of interactions between those users. Even for smaller networks, one may require real-time analysis of the networks, which make using the entire dataset for running analysis intractable. To address the computational complexity of running analysis on large networks we have introduced a sampling technique, which helps reduce the computational cost of running analysis on large and medium size networks. In this section, we introduce this sampling module and explore the impact of using sampling on various statistical properties of the networks as well as the corresponding graph embeddings. Some asymptotic analysis of the effect of sampling was presented in~\cite{touwen2024whole}.

\subsubsection{Sampling Technique and Metrics}

Here, we introduce the sampling technique used in \textbf{NEExT} and define metrics we use in our experiments to analyze the impact of sampling on graph embeddings. As highlighted previously, the \textbf{NEExT} framework works in two steps. First, we compute a set of node features on all nodes of every graph in the graph collection. Secondly, we use the computed node features to build embedding vectors for each graph in the graph collection. The major contributor to the computational cost is the \textit{first} step of this process, which computes node features on every node of every graph. The embedding generation is computationally less costly since we use an efficient algorithm, the approximate Wasserstein vectorization algorithm.

A natural way to reduce the computational cost is to compute node features on a subset of nodes in each graph, instead of all nodes. Since the goal is to compute node features only on a subset of nodes, sampling nodes based on certain node features (such as node degree) would not be feasible. This is because it would require us compute node degree on all nodes, which defeats the purpose. More importantly, the Wasserstein distance treats all nodes equally in the associated linear optimal transport task. Therefore, we sample nodes for each graph uniformly at random, with sampling rate $r$. Once we have a random subset of nodes sampled for each graph, we compute node features on the sampled nodes and build graph embeddings using those features. It is important to note that sampling does not impact the structure of the graphs themselves, but only what nodes features are computed for. This means that sampling does not reduce \textit{space complexity} of the process, but only the \textit{time complexity}.

Although sampling reduces computational cost of building and analyzing graph embeddings, it is important to consider the effect of sampling on the quality of embeddings generated and their effects on any downstream tasks such as graph classification done using computed graph embeddings. To explore the impact of sampling on the above, we perform two types of experiments:
\begin{itemize}
    \item \textbf{Experiment 5:} Comparing the similarity between embeddings generated using different sampling rates.
    \item \textbf{Experiment 6:} Analyzing the accuracy of classifiers trains on embeddings build using different sampling rates.
\end{itemize}
The results and analysis of the above experiments are presented in the following subsections.

\subsubsection{Experiment 5 -- Effect of Sampling on Embedding Quality}

We start our analysis by exploring the impact of sampling on the quality of embeddings. First, we note that in the following experiments sampling is measured and defined as $(1-r)$, where $r$ is the sample rate. It is a metric that ranges between 0 and 1 and measures the fraction of nodes removed from each graph. For example, $1-r=0$, means that we are using all the nodes in the graphs to compute node features and embedding vectors. Secondly, we define the \textbf{Embedding Similarity Score} in the following way:
\begin{itemize}
    \item We compute reference graph embeddings for all the graphs in the graph collection using all the nodes. Lets call this embedding space $E_i$
    \item Then, for a given sample rate \textbf{r}, we compute node features and embeddings for all the graphs in the graph collection using the sampled nodes. Lets call this embedding space $E_j$.
    \item The \textbf{Embedding Similarity Score} tells us how similar, as measured using \textbf{Wasserstein distance}, are the distances between various graphs in embedding space $E_i$ to distances in embedding space $E_j$.
    \item Distances in each embedding space is measured by selecting a reference graph $G_o$, and measuring the similarity distance (measured using \textbf{Wasserstein distance}) between the reference graph $G_o$ and a randomly selected set of graphs within the collection $s_0 = \text{Sim}(G_o, G_k)$. This provides us with a list of similarity measures in each embedding space: $S_{E_i} = [s_0, s_1, ..., s_k]$.
    \item The \textbf{Embedding Similarity Score} is then defined as $S = \text{Sim}(S_{E_i}, S_{E_j})$.
\end{itemize}

We have chosen this similarity score, as it provides us with a natural way of comparing two embedding spaces. The \textbf{Embedding Similarity Score} ranges between 0 and 1. This metric is measured between embedding spaces build using sampling rate $r$ and the original embedding space built using all the nodes ($1-r=0$).

\begin{figure}[ht]
  \centering
  \includegraphics[scale=0.3]{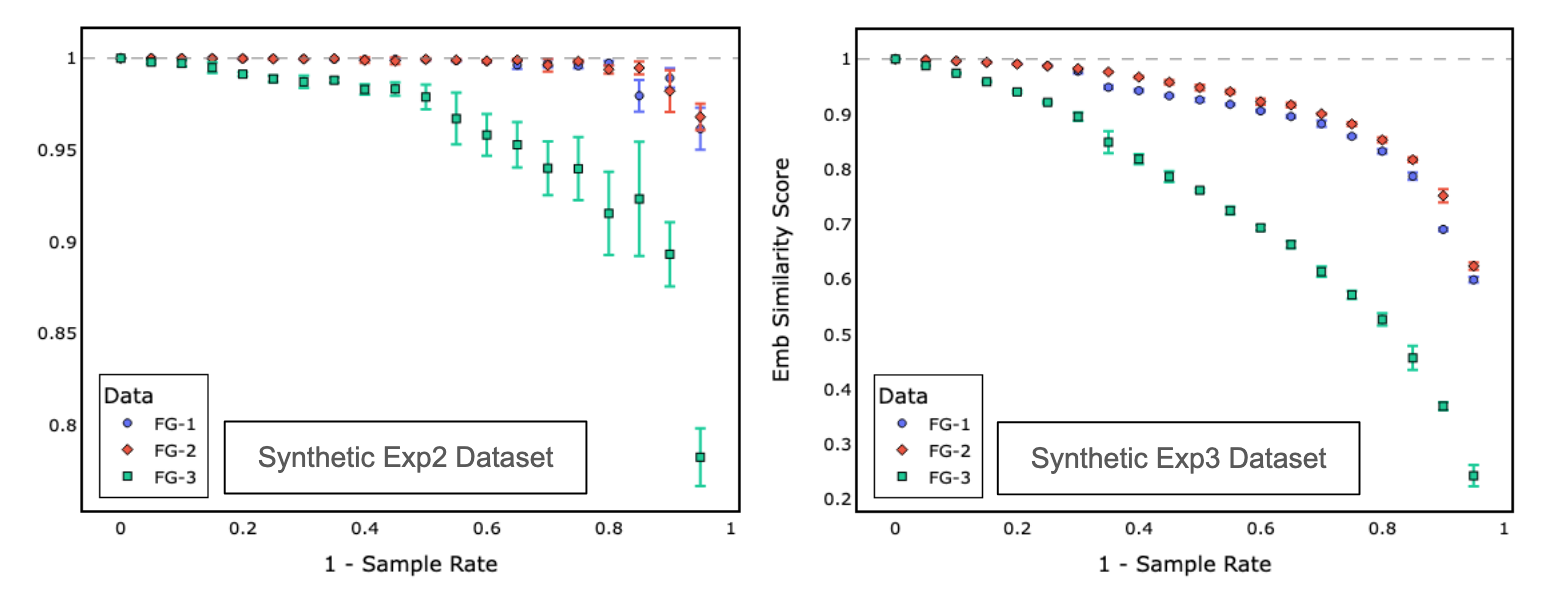}
  \caption{Plot of Embedding Similarity score as a function of $1-r$ for two synthetic datasets. Dataset \textbf{Exp2} and \textbf{Exp3} are datasets used in Subsections~\ref{ssec:exp_synthetic_exp2} and~\ref{ssec:exp_synthetic_exp3}, respectively.}
  \label{fig:Synthetic_sampling_Analysis_Feat_Groups}
\end{figure}

First, we measure \textbf{Embedding Similarity Score} for two of the synthetic datasets \textbf{Exp2} and \textbf{Exp3}, introduced in Subsections~\ref{ssec:exp_synthetic_exp2} and~\ref{ssec:exp_synthetic_exp3}, respectively. The results of these experiments are shown in Figure~\ref{fig:Synthetic_sampling_Analysis_Feat_Groups}. Here, for any given sample rate, we compute a group of node features, labeled as FG-1 to FG-3, and use those features to build graph embeddings for each graph in the collection. Mapping between feature groups and features are given in Table~\ref{table:embedding_similarity_feature_groups}.

\begin{table}[ht]
  \begin{center}
    \begin{tabular}{|c|p{0.6\linewidth}|}
      \hline
      \textbf{Feature Group} & \textbf{Node Features}\\
      \hline
      \hline
      FG-1 & All features\\
      FG-2 & \textbf{PageRank}, \textbf{Degree Centrality}, \textbf{Closeness Centrality}, \textbf{Load Centrality}, \textbf{Eigenvector Centrality} \\
      FG-3 & \textbf{LSME}, \textbf{Expansion} \\
      \hline
      \hline
    \end{tabular}
    \caption{Definition of feature groups used for the experiment shown in Figure~\ref{fig:Synthetic_sampling_Analysis_Feat_Groups}.}
    \label{table:embedding_similarity_feature_groups}
  \end{center}
\end{table}

The overall qualitative trend we see in these experiments follows the expected logic that as we build embeddings using smaller set of sampled nodes, the quality of the embeddings decreases. Here, as we mentioned before, the quality is measured by computing the similarity between the embeddings generated some sample rate $r$ and embeddings generated using all the nodes in the graphs.

Looking closer at the data, we see that quality of some feature groups decrease faster than others. For example, in both experiments (\textbf{Exp2} and \textbf{Exp3}), sampling has a more significant impact on embeddings generated using feature group FG-3 (\textbf{LSME} and \textbf{Expansion}). This effect is a lot more noticeable in \textbf{Exp3}. The results shown in these experiments hint at two important findings. First, the effect of sampling on the quality of embeddings generated will depend on the underlying dataset. For example, as we show in the right chart in Figure~\ref{fig:Synthetic_sampling_Analysis_Feat_Groups}, using feature groups FG-1 and FG-2, one could remove up to $75\%$ of the nodes, without noticeable impact on the quality of the embeddings. The second finding highlights the fact that each feature, as shown by comparing different feature groups, will be impacted differently by the sampling rate. Lastly, it is important to consider the fact that there are more features in FG-1 than there are in FG-2, and similarly, there are more features in FG-2 than there are in FG-3. This will have an impact on the quality of the embeddings. What our results show is that if sampling is used to enhance the computational efficiency of the process, it is important to consider the impact of sampling on each feature (feature group) and it may be more efficient to enhance the quality by adding more features than reducing sampling rate.

\subsubsection{Experiment 6 -- Effect of Sampling on Models}

In the above experiments, we have shown that the quality of generated embedding vectors decreases as we remove more nodes from the underlying graphs through the sampling process. However, most practitioners are interested in using the generated embedding vectors to build downstream machine learning models such as graph classifiers. Therefore, it is important to investigate the impact of sampling on the quality of such downstream models. To answer this question, we performed experiments that analyze the impact of sampling on graph classification for graphs in the real-world dataset. Similar to the previous section, we use $1-r$ as the sampling parameter to adjust the fraction of nodes sampled in each experiment. The following outlines the experimentation process:

\begin{itemize}
    \item For each dataset, we choose a sampling rate ($1-r$) and sample nodes from each graph in the collection accordingly.
    \item We compute all node features (as defined in Section~\ref{sec:tool}) for the sampled nodes, for each graph. This results in a $32$ dimensional feature vector.
    \item The node features are used to compute $32$ dimensional embedding vectors using the approximate Wasserstein vectorization algorithm.
    \item The embeddings are used as features in building graph classification models, where we train $50$ XGBoost models with 70/30 train/test split.
    \item The accuracy of models are computed.
\end{itemize}

\begin{figure}[ht]
  \centering
  \includegraphics[scale=0.56]{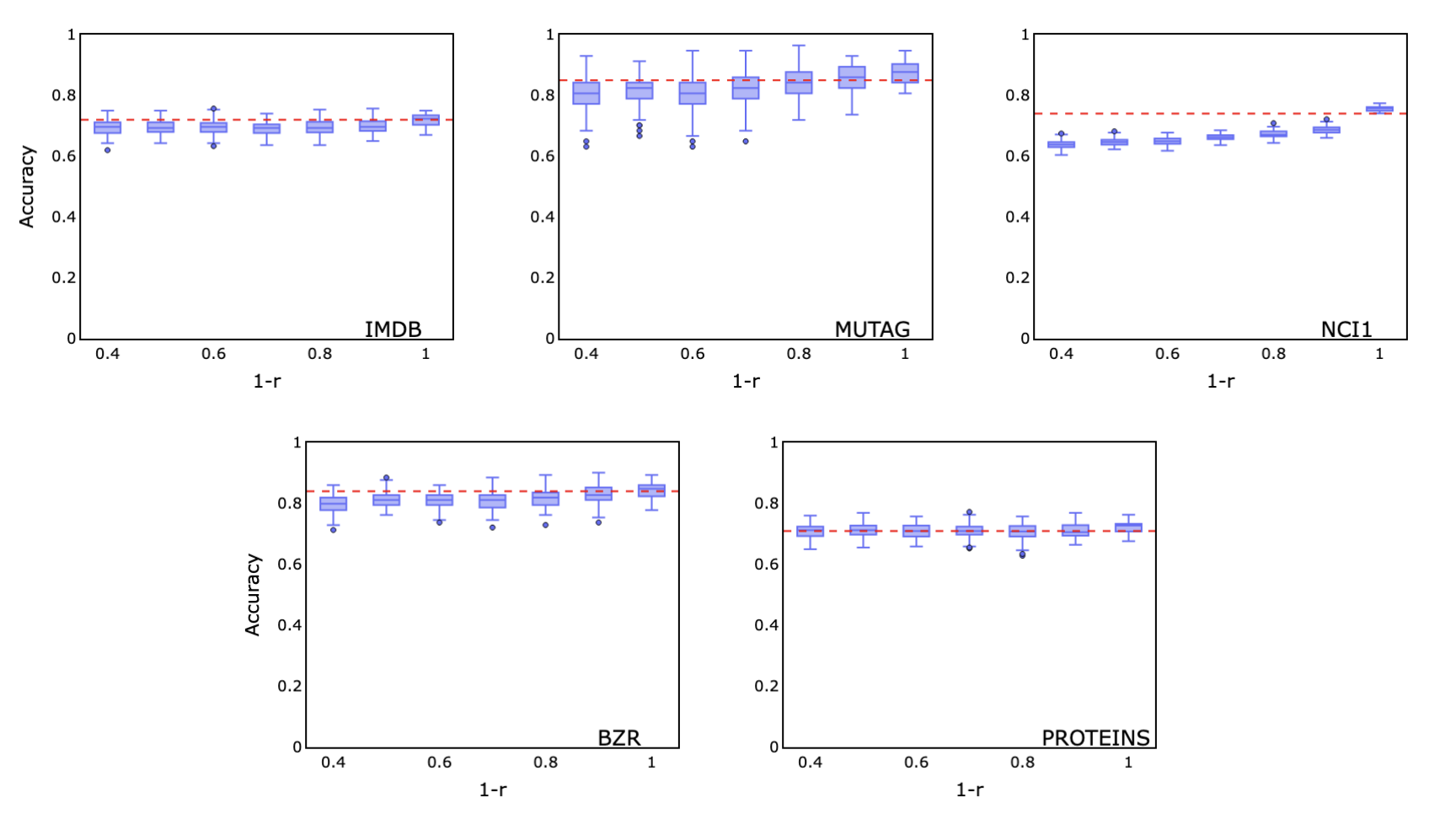}
  \caption{Accuracy of classification models built using embeddings at different sample rates ($1-r$), for all real-world datasets.}
  \label{fig:Accuracy_vs_SampleRate_MUTAG_Synthetic_Exp3}
\end{figure}

The results of the above experiments are shown in Figure~\ref{fig:Accuracy_vs_SampleRate_MUTAG_Synthetic_Exp3}. Although we know that sampling rate impacts the quality of the underlying embeddings, the results in Figure~\ref{fig:Accuracy_vs_SampleRate_MUTAG_Synthetic_Exp3} show that in these scenarios, the quality of the classifiers built on those embeddings is not impacted. This has to do with the fact that classifiers look for dividing boundaries between classes, and as long as this boundary can be found, a classifier can perform well. In addition, the approximate Wasserstein vectorization algorithm measures the distance based on the centre of the mass of the distributions, which is not impacted as greatly by the removal of nodes. This helps dampen the effect of sampling on the quality of the generated models. 

\begin{figure}[ht]
  \centering
  \includegraphics[scale=0.45]{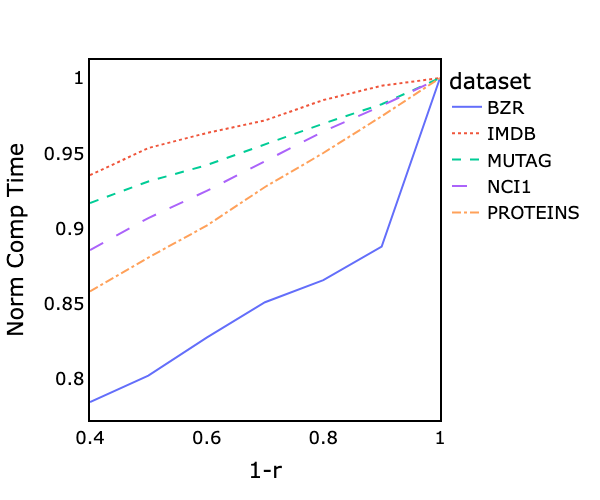}
  \caption{Normalized computational time for running experiments for each real-world dataset as a function of sampling rate.}
  \label{fig:normalized_computational_cost_sampling_exp}
\end{figure}

In Figure~\ref{fig:normalized_computational_cost_sampling_exp}, we show the normalized computation time for each experiment. We normalize the average computation time with respect to the case where all nodes are used for the experiment. As we can see, for example, sampling only 50\% of the nodes results in a computation time reduction between 5-20\%. The primary reason why sampling does not have a more significant impact on computation time is that node features such as \textit{PageRank} or some \textbf{Centrality} measures are global measures, which means that we need to compute them across all nodes. Lastly, as we can see in Figure~\ref{fig:normalized_computational_cost_sampling_exp}, computation time of datasets is impacted differently, depending on the structure of the network. For example, \textbf{BZR} has the largest time reduction as a function of sampling rate, while \textbf{IMDB} has the smallest. This points to the fact that some graphs could have a small set of computationally expensive nodes (perhaps outliers), which contribute to the computational cost while not impacting the quality of the resulting classifier models. We conclude this section by highlighting the fact that one could reduce the computational cost of running NEExT on large datasets by sampling only 50-60\% of nodes and arrive at roughly equal quality models. However, the exact reduction in the computational cost will depend on the global versus local nature of the selected node features and the structure and connectivity of the underlying graph.

\section{Conclusion}\label{sec:conclusion}

In this paper, we introduced \textbf{NEExT} (\textbf{N}etwork \textbf{E}mbedding \textbf{Ex}ploration \textbf{T}ool), a Python framework for exploring and analyzing collections of graphs. Using synthetic and real-world datasets, we performed experiments to highlight various capabilities of our framework, such as building graph embeddings and performing feature importance analysis. We explored the computational efficiency of our framework through graph \textbf{sampling} and computationally efficient graph embedding techniques such as the \textbf{Approximate Wasserstein} algorithm. We also showed that the models and embeddings generated using our framework are explainable, as the underlying features are built using known and commonly used node-features. We compared the performance classifier models created using \textbf{NEExT} to other cutting edge techniques across five real-world datasets, and showed the models generated by \textbf{NEExT} are comparable in performance, while also being explainable. Finally, we want to point out that although \textbf{NEExT} is designed to work with a collection of graphs, one could also use our framework on single networks, by composing the collection of communities or ego-nets around various nodes of a single graph. We hope that \textbf{NEExT} is a useful contribution to the network science community.

\bibliographystyle{splncs04}
\bibliography{bibliography}

\end{document}